\documentclass[10pt,twocolumn,letterpaper]{article}

\usepackage[pagenumbers]{cvpr} 

\usepackage{graphicx, amsmath, amssymb, caption, subcaption, textpos}
\usepackage[table]{xcolor}
\usepackage[british, english, american]{babel}
\usepackage[pagebackref=false, breaklinks=true, letterpaper=true, colorlinks,
            citecolor=citecolor, linkcolor=linkcolor, bookmarks=false]{hyperref}
\definecolor{citecolor}{HTML}{0071BC}
\definecolor{linkcolor}{HTML}{ED1C24}

\definecolor{baselinecolor}{gray}{.92}
\newcommand{\baseline}[1]{\cellcolor{baselinecolor}{#1}}

\newcommand{\lr}{\emph{lr}\xspace}
\newcommand{\wtd}{\emph{wd}\xspace}
\newcommand{\drp}{\emph{dp}\xspace}
\newcommand{\expnum}[2]{{#1}\mathrm{e}{#2}}
\newcommand{\dalle}{\texttt{DALL$\cdot$E}\xspace}
\newcommand{\boxAP}{AP$^\text{box}$\xspace}

\newcommand{\app}{\raise.17ex\hbox{$\scriptstyle\sim$}}
\newlength\savewidth\newcommand\shline{\noalign{\global\savewidth\arrayrulewidth
  \global\arrayrulewidth 1pt}\hline\noalign{\global\arrayrulewidth\savewidth}}
\newcommand{\tablestyle}[2]{\setlength{\tabcolsep}{#1}\renewcommand{\arraystretch}{#2}\centering\footnotesize}
\makeatletter\renewcommand\paragraph{\@startsection{paragraph}{4}{\z@}
  {.5em \@plus1ex \@minus.2ex}{-.5em}{\normalfont\normalsize\bfseries}}\makeatother
\newcolumntype{x}[1]{>{\centering\arraybackslash}p{#1pt}}
\newcolumntype{y}[1]{>{\raggedright\arraybackslash}p{#1pt}}
\newcolumntype{z}[1]{>{\raggedleft\arraybackslash}p{#1pt}}

\usepackage{algorithm}
\usepackage{listings}
\usepackage{etoolbox}
\makeatletter
\AfterEndEnvironment{algorithm}{\let\@algcomment\relax}
\AtEndEnvironment{algorithm}{\kern2pt\hrule\relax\vskip3pt\@algcomment}
\let\@algcomment\relax
\newcommand\algcomment[1]{\def\@algcomment{\footnotesize#1}}
\renewcommand\fs@ruled{\def\@fs@cfont{\bfseries}\let\@fs@capt\floatc@ruled
  \def\@fs@pre{\hrule height.8pt depth0pt \kern2pt}%
  \def\@fs@post{}%
  \def\@fs@mid{\kern2pt\hrule\kern2pt}%
  \let\@fs@iftopcapt\iftrue}
\makeatother

\def\cvprPaperID{****}
\def\confName{****}
\def\confYear{****}

\begin{document}

\title{Benchmarking Detection Transfer Learning with Vision Transformers}

\author{%
Yanghao Li \quad Saining Xie \quad Xinlei Chen \quad Piotr Doll\'ar \quad Kaiming He \quad Ross Girshick\\[3mm]
Facebook AI Research (FAIR)}
\maketitle

\begin{abstract}
Object detection is a central downstream task used to test if pre-trained network parameters confer benefits, such as improved accuracy or training speed. The complexity of object detection methods can make this benchmarking non-trivial when new architectures, such as Vision Transformer (ViT) models, arrive. These difficulties (\eg, architectural incompatibility, slow training, high memory consumption, unknown training formulae, \etc) have prevented recent studies from benchmarking detection transfer learning with standard ViT models. In this paper, we present training techniques that overcome these challenges, enabling the use of standard ViT models as the backbone of Mask R-CNN. These tools facilitate the primary goal of our study: we compare five ViT initializations, including recent state-of-the-art self-supervised learning methods, supervised initialization, and a strong random initialization baseline. Our results show that recent masking-based unsupervised learning methods may, for the first time, provide convincing transfer learning improvements on COCO, increasing \boxAP up to 4\% (absolute) over supervised and prior self-supervised pre-training methods. Moreover, these masking-based initializations scale better, with the improvement growing as model size increases.
\end{abstract}


\section{Introduction}
\label{sec:intro}
Unsupervised/self-supervised deep learning is commonly used as a \emph{pre-training} step that initializes model parameters before they are \emph{transferred} to a downstream task, such as image classification or object detection, for \emph{fine-tuning}. The utility of an unsupervised learning algorithm is judged by downstream task metrics (\eg accuracy, convergence speed, \etc) in comparison to baselines, such as \emph{supervised} pre-training or \emph{no} pre-training at all, \ie, random initialization (often called training ``from scratch'').

Unsupervised deep learning in computer vision typically uses standard convolutional network (CNN) models~\cite{LeCun1989}, such as ResNets~\cite{He2016}. Transferring these models is relatively straightforward because CNNs are in widespread use in most downstream tasks, and thus benchmarking protocols are easy to define and baselines are plentiful (\eg~\cite{He2020}). In other words, unsupervised learning with CNNs produces a plug-and-play parameter initialization.

We are now witnessing the growth of unsupervised learning with \emph{Vision Transformer} (ViT) models~\cite{Dosovitskiy2021}, and while the high-level transfer learning methodology remains the same, the low-level details and baselines for some important downstream tasks have not been established. Notably, object detection, which has played a central role in the study of transfer learning over the last decade (\eg,~\cite{Sermanet2013,Girshick2014,Doersch2015,He2020}), was not explored in the pioneering work on ViT training~\cite{Dosovitskiy2021,Chen2021a,Caron2021}---supervised or unsupervised---due to the challenges (described shortly) of integrating ViTs into common detection models, like Mask R-CNN~\cite{He2017}.

To bridge this gap, this paper establishes a transfer learning protocol for evaluating ViT models on object detection and instance segmentation using the COCO dataset~\cite{Lin2014} and the Mask R-CNN framework. We focus on standard ViT models, with minimal modifications, as defined in the original ViT paper~\cite{Dosovitskiy2021}, because we expect this architecture will remain popular in unsupervised learning work over the next few years due to its simplicity and flexibility when exploring new techniques, \eg, masking-based methods~\cite{Bao2021,he2021mae}.

Establishing object detection baselines for ViT is challenging due to technical obstacles that include mitigating ViT's large memory requirements when processing detection-sized inputs (\eg, \app 20$\times$ more patches than in pre-training), architectural incompatibilities (\eg, single-scale ViT \vs a multi-scale detector), and developing effective training formulae (\ie, learning schedules, regularization and data augmentation methods, \etc) for numerous pre-trained initializations, as well as random initialization. We overcome these obstacles and present strong ViT-based Mask R-CNN baselines on COCO when initializing ViT from-scratch~\cite{He2019}, with pre-trained ImageNet~\cite{Deng2009} supervision, and with unsupervised pre-training using recent methods like MoCo v3~\cite{Chen2021a}, BEiT~\cite{Bao2021}, and MAE~\cite{he2021mae}.

\begin{figure}[t]\centering
\includegraphics[width=.99\linewidth,trim={0 0 0 0},clip]{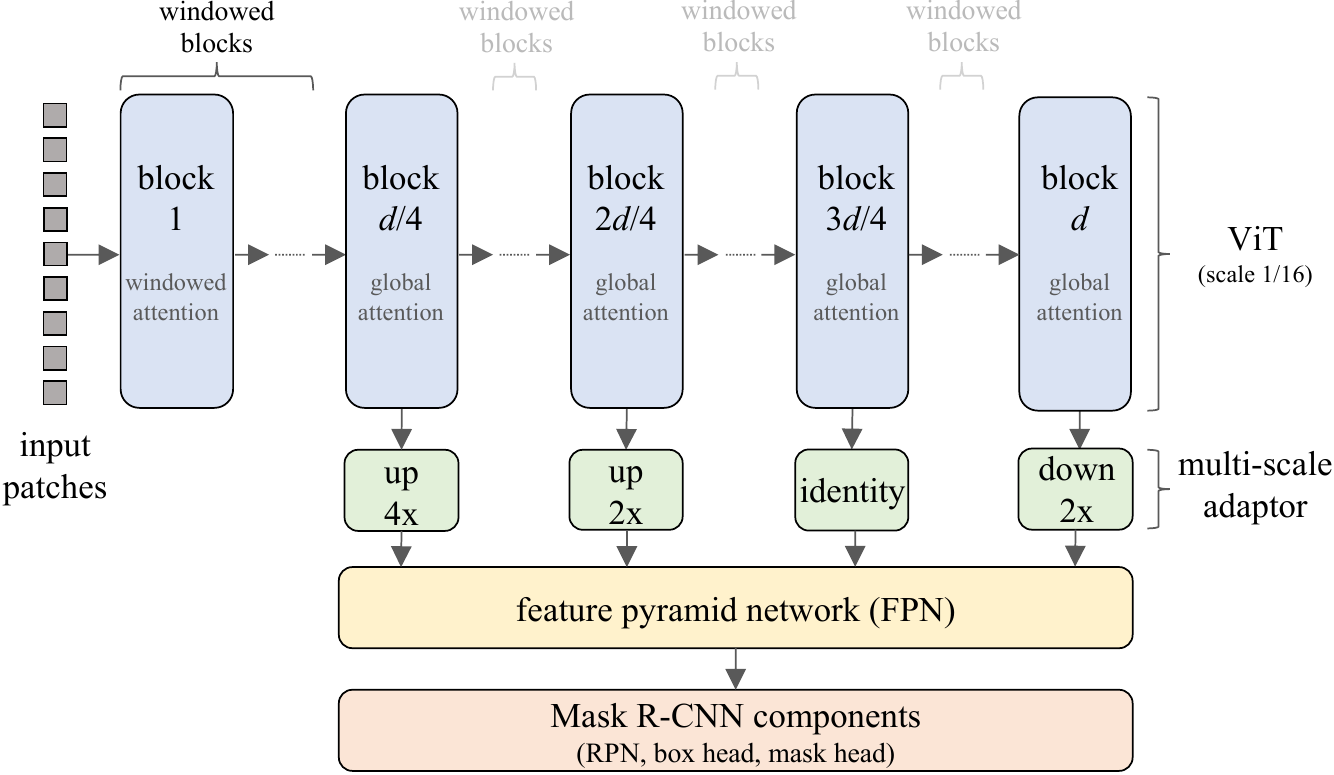}
\vspace{-.5em}
\caption{\textbf{ViT-based Mask R-CNN.} In \S\ref{sec:approach} we describe how a standard ViT model can be used effectively as the backbone in Mask R-CNN. To save time and memory, we modify the ViT to use non-overlapping windowed attention in all but four of its Transformer blocks, spaced at an interval of $d/4$, where $d$ is the total number of blocks (blue)~\cite{mvitv2}. To adapt the single-scale ViT to the multi-scale FPN (yellow), we make use of upsampling and downsampling modules (green)~\cite{el2021xcit}. The rest of the system (light red) uses upgraded, but standard, Mask R-CNN components.
\label{fig:arch}
}
\vspace{-3mm}
\end{figure}

Looking beyond ViT, we hope our practices and observations will serve as a blueprint for future work comparing pre-training methods for more advanced ViT derivatives, like Swin~\cite{liu2021swin} and MViT~\cite{fan2021multiscale}. To facilitate community development we will release code in Detectron2~\cite{Wu2019}.


\section{Approach}
\label{sec:approach}

We select the Mask R-CNN~\cite{He2017} framework due to its ubiquitous presence in object detection and transfer learning research. Mask R-CNN is the foundation of higher complexity/higher performing systems, such as Cascade \mbox{R-CNN}~\cite{Cai2018} and HTC/HTC++~\cite{chen2019hybrid,liu2021swin}, which may improve upon the results presented here at the cost of additional complexity that is orthogonal to the goal of benchmarking transfer learning. Our choice attempts to balance (relative) simplicity \vs complexity while providing compelling, even though not entirely state-of-the-art, results.

We configure Mask R-CNN with a number of upgraded modules (described in \S\ref{sec:modules}) and training procedures (described in \S\ref{sec:formula}) relative to the original publication. These upgrades, developed primarily in~\cite{Wu2018,He2019,ghiasi2021simple}, allow the model to be trained effectively from random initialization, thus enabling a meaningful from-scratch baseline. Next, we will discuss how the backbone, which would typically be a ResNet, can be replaced with a Vision Transformer.

\subsection{ViT Backbone}
In this section we address two technical obstacles when using ViT as the backbone in Mask R-CNN: (1) how to adapt it to work with a feature pyramid network (FPN)~\cite{Lin2017} and (2) how to reduce its memory footprint and runtime to make benchmarking large ViT backbones tractable.

\paragraph{FPN Compatibility.} Mask R-CNN can work with a backbone that either produces a single-scale feature map or feature maps at multiple scales that can be input into an FPN. Since FPN typically provides better detection results with minimal time and memory overhead, we adopt it.

However, using FPN presents a problem because ViT produces feature maps at a single scale (\eg, $1/16$th), in contrast to the multi-scale feature maps produced by typical CNNs.\footnote{We view the natural 2D spatial arrangement of intermediate ViT patch embeddings as a standard 2D feature map.} To address this discrepancy, we employ a simple technique from~\cite{el2021xcit} (used for the single-scale XCiT backbone) to either upsample or downsample intermediate ViT feature maps by placing four resolution-modifying modules at equally spaced intervals of $d/4$ transformer blocks, where $d$ is the total number of blocks. See Figure~\ref{fig:arch} (green blocks).

The first of these modules upsamples the feature map by a factor of $4$ using a stride-two $2\times 2$ transposed convolution, followed by group normalization~\cite{Wu2018} and GeLU~\cite{Hendrycks2016}, and finally another stride-two $2\times 2$ transposed convolution. The next $d/4$th block's output is upsampled by $2\times$ using a single stride-two $2\times 2$ transposed convolution (without normalization and non-linearity). The next $d/4$th block's output is taken as is and the final ViT block's output is downsampled by a factor of two using stride-two $2\times 2$ max pooling. Each of these modules preserves the ViT's embedding/channel dimension. Assuming a patch size of $16$, these modules produce feature maps with strides of $4$, $8$, $16$, and $32$ pixels, \wrt the input image, that are ready to input into an FPN.

We note that recent work, such as Swin~\cite{liu2021swin} and MViT~\cite{fan2021multiscale}, address the single \vs multi-scale feature map problem by modifying the core ViT architecture (in pre-training) so it is inherently multi-scale. This is an important direction, but it also complicates the simple ViT design and may impede the exploration of new unsupervised learning directions, such as methods that sparsely process unmasked patches~\cite{he2021mae}. Therefore, we focus on \emph{external} additions to ViTs that allow them to integrate into multi-scale detection systems. We also note that Beal \etal~\cite{beal2020toward} integrate standard ViT models with Faster R-CNN~\cite{Ren2017}, but report substantially lower \boxAP compared to our results ($>$10 points lower), which suggests that our design is highly effective.

\paragraph{Reducing Memory and Time Complexity.} Using ViT as a backbone in Mask R-CNN introduces memory and runtime challenges. Each self-attention operation in ViT takes $O(h^2w^2)$ space and time for an image tiled (or ``patchified'') into $h \times w$ non-overlapping patches~\cite{Vaswani2017}.

During pre-training, this complexity is manageable as $h = w = 14$ is a typical setting (a $224\times 224$ pixel image patchified into $16\times 16$ pixel patches). In object detection, a standard image size is $1024\times 1024$---\emph{approximately $21\times$ more pixels and patches}. This higher resolution is needed in order to detect relatively small objects as well as larger ones. Due to the quadratic complexity of self-attention, even the ``base'' size ViT-B may consume \app $20$--$30$GB of GPU memory when used in Mask R-CNN with a \emph{single}-image minibatch and \emph{half}-precision floating point numbers.

To reduce space and time complexity we use restricted (or ``windowed'') self-attention~\cite{Vaswani2017}, which saves both space and time by replacing global computation with local computation. We partition the $h\times w$ patchified image into $r\times r$ patch \emph{non-overlapping} windows and compute self-attention \emph{independently} within each of these windows. This windowed self-attention has $O(r^2hw)$ space and time complexity (from $O(r^4)$ per-window complexity and $h/r\times w/r$ windows). We set $r$ to the global self-attention size used in pre-training (\eg, $r=14$ is typical).

A drawback of windowed self-attention is that the backbone does not integrate information across windows. Therefore we adopt the hybrid approach from~\cite{mvitv2} that includes four global self-attention blocks placed evenly at each $d/4$th block (these coincide with the up-/downsampling locations used for FPN integration; see Figure~\ref{fig:arch}).

\subsection{Upgraded Modules}
\label{sec:modules}
Relative to the original Mask R-CNN in~\cite{He2017}, we modernize several of its modules. Concisely, the modifications include: (1) following the convolutions in FPN with batch normalization (BN)~\cite{Ioffe2015}, (2) using two convolutional layers in the region proposal network (RPN)~\cite{Ren2015} instead of one, (3) using four convolutional layers with BN followed by one linear layer for the region-of-interest (RoI) classification and box regression head~\cite{Wu2018} instead of a two-layer MLP without normalization, (4) and following the convolutions in the standard mask head with BN. Wherever BN is applied, we use \emph{synchronous} BN across all GPUs. These upgrades are implemented in the Detectron2 model zoo.\footnote{\scriptsize\url{https://github.com/facebookresearch/detectron2/blob/main/MODEL_ZOO.md\#new-baselines-using-large-scale-jitter-and-longer-training-schedule}}

\subsection{Training Formula}
\label{sec:formula}
We adopt an upgraded training formula compared to the original Mask R-CNN. This formula was developed in~\cite{He2019}, which demonstrated good from-scratch performance when training with normalization layers and for long enough, and~\cite{ghiasi2021simple}, which demonstrated that a simple data augmentation method called \emph{large-scale jitter} (LSJ) is effective at preventing overfitting and improves results when models are trained for \emph{very} long schedules (\eg, 400 epochs).

We aim to keep the number of hyperparameters low and therefore resist adopting additional data augmentation and regularization techniques. However, we found that drop path regularization~\cite{Larsson2016,Huang2016} is highly effective for ViT backbones and therefore we include it (\eg, it improves from-scratch training by up to 2 \boxAP).

In summary, we train all models with the same simple formula: LSJ ($1024\times 1024$ resolution, scale range $[0.1, 2.0]$), AdamW~\cite{Loshchilov2019} ($\beta_1, \beta_2 = 0.9, 0.999$) with half-period cosine learning rate decay, linear warmup~\cite{Goyal2017} for $0.25$ epochs, and drop path regularization. When using a pre-trained initialization, we fine-tune Mask R-CNN for up to 100 epochs. When training from scratch, we consider schedules of up to 400 epochs since convergence is slower than when using pre-training. We distribute training over 32 or 64 GPUs (NVIDIA V100-32GB) and always use a minibatch size of 64 images. We use PyTorch's automatic mixed precision. Additional hyperparameters are tuned by the consistent application of a protocol, describe next.

\subsection{Hyperparameter Tuning Protocol}
To adapt the training formula to each model, we tune three hyperparameters---learning rate (\lr), weight decay (\wtd), and drop path rate (\drp)---while keeping all others the same for all models. We conducted pilot experiments using ViT-B pre-trained with MoCo v3 to estimate reasonable hyperparameter ranges. Based on these estimates we established the following tuning protocol:

\textbf{(1)} For each initialization (from-scratch, supervised, \etc), we fix \drp at $0.0$ and perform a grid search over \lr and \wtd using ViT-B and a 25 epoch schedule (or 100 epochs when initializing from scratch). We center a $3\times 3$ grid at \lr, \wtd $= \expnum{1.6}{-4}$, $0.1$ and use doubled and halved values around the center. If a local optimum is not found (\ie the best value is a boundary value), we expand the search.

\textbf{(2)} For ViT-B, we select \drp from $\{0.0, 0.1, 0.2, 0.3\}$ using a 50 epoch schedule for pre-trained initializations. The shorter 25 epoch schedule was unreliable and 100 epochs was deemed impractical. For random initialization we're forced to use 100 epochs due to slow convergence. We found that \drp $=0.1$ is optimal for all initializations.

\textbf{(3)} For ViT-L, we adopt the optimal \lr and \wtd from ViT-B (searching with ViT-L is impractical) and find \drp $=0.3$ is best using the same procedure as for ViT-B.

\paragraph{Limitations.} The procedure above takes practical shortcuts to reduce the full hyperparameter tuning space. In particular, \lr and \wtd are optimized separately from \drp, thus the combination may be suboptimal. Further, we only tune \lr and \wtd using ViT-B, therefore the choice may be suboptimal for ViT-L. We also tune \lr and \wtd using a schedule that is $4\times$ shorter than the longest schedule we eventually train at, which again may be suboptimal. Given these limitations we aim to avoid biasing results by applying the \emph{same} tuning protocol to all initializations.

Finally, we note that we tune \lr, \wtd, and \drp on the COCO 2017 \texttt{val} split and report results on the same split. While technically not an ML best-practice, a multitude of comparisons on COCO \texttt{val} \vs \texttt{test-dev} results over many years demonstrate that overfitting in not a concern for this kind of low-degree-of-freedom hyperparameter tuning.\footnote{\Eg, Table 2 in~\cite{liu2021swin} (version 1) shows that \texttt{text-dev} \boxAP is systematically higher than \texttt{val} \boxAP in seven system-level comparisons.}

\subsection{Additional Implementation Details}
Images are padded during training and inference to form a $1024\times 1024$ resolution input. During training, padding is necessary for batching. During (unbatched) inference, the input only needs to be a multiple of the ViT patch size on each side, which is possibly less than $1024$ on one side. However, we found that such reduced padding performs worse (\eg, decrease of \app 0.5--1 \boxAP) than padding to the same resolution used during training, likely due to ViT's use of positional information. Therefore, we use a $1024\times 1024$ resolution input at inference time, even though the extra padding slows inference time by \app 30\% on average.

\section{Initialization Methods}
\label{sec:init_methods}

We compare five initialization methods, which we briefly summarize below.

\textit{Random:} All network weights are randomly initialized and no pre-training is used. The ViT backbone initialization follows the code of~\cite{Bao2021} and the Mask R-CNN initialization uses the defaults in Detectron2~\cite{Wu2019}.

\textit{Supervised:} The ViT backbone is pre-trained for supervised classification using ImageNet-1k images \emph{and} labels. We use the DeiT released weights~\cite{Touvron2020} for ViT-B and the ViT-L weights from~\cite{he2021mae}, which uses an even stronger training formula than DeiT to avoid overfitting (moreover, the DeiT release does not include ViT-L). ViT-B and ViT-L were pre-trained for 300 and 200 epochs, respectively.

\textit{MoCo v3:} We use the unsupervised ImageNet-1k pre-trained ViT-B and ViT-L weights from the authors of~\cite{Chen2021a} (\mbox{ViT-B} is public; ViT-L was provided via private communication). These models were pre-trained for 300 epochs.

\textit{BEiT:} Since ImageNet-1k pre-trained weights are not available, we use the official BEiT code release~\cite{Bao2021} to train ViT-B and ViT-L ourselves for 800 epochs (the default training length used in~\cite{Bao2021}) on unsupervised ImageNet-1k.

\textit{MAE:} We use the ViT-B and ViT-L weights pre-trained on unsupervised ImageNet-1k from the authors of~\cite{he2021mae}. These models were pre-trained for 1600 epochs using \emph{normalized} pixels as the target.

\subsection{Nuisance Factors in Pre-training}
\label{sec:nuisance}
We attempt to make comparisons as equally matched as possible, yet there are pre-training nuisance factors, listed below, that differ across methods.

\textbf{(1)} Different pre-training methods may use different numbers of epochs. We adopt the default number of pre-training epochs from the respective papers. While these values may not \emph{appear} comparable, the reality is unclear: not all methods may benefit equally from longer training and not all methods have the same per-epoch training cost (\eg, BEiT uses roughly $3\times$ more flops than MAE).

\textbf{(2)}
BEiT uses learned relative position biases that are added to the self-attention logits~\cite{raffel2019exploring} in each block, instead of the absolute position embeddings used by the other methods. To account for this, albeit imperfectly, we include \emph{both} relative position biases and absolute position embeddings in all detection models regardless of their use in pre-training. For BEiT, we transfer the pre-trained biases and randomly initialize the absolute position embeddings. For all other methods, we zero-initialize the relative position biases and transfer the pre-trained absolute position embeddings. Relative position biases are shared across windowed attention blocks and (separately) shared across global attention blocks. When there is a spatial dimension mismatch between pre-training and fine-tuning, we resize the pre-trained parameters to the required fine-tuning resolution.

\textbf{(3)} BEiT makes use of layer scale~\cite{touvron2021going} in pre-training, while the other methods do not. During fine-tuning, the BEiT-initialized model must also be parameterized to use layer scale with the pre-trained layer scaling parameters initialized from the pre-trained model. All other models do not use layer scale in pre-training or in fine-tuning.

\textbf{(4)} We try to standardize pre-training data to ImageNet-1k, however BEiT uses the \dalle~\cite{Ramesh2021} discrete VAE (dVAE), which was trained on \app 250 million proprietary and undisclosed images, as an image tokenizer. The impact of this additional training data is not fully understood.

\section{Experiments and Analysis}
\label{sec:experiments}

\subsection{Comparing Initializations}

\begin{table}
\tablestyle{6pt}{1.1}
\begin{tabular}{@{}lccccc@{}}
& pre-training & \multicolumn{2}{c}{AP$^\text{box}$} & \multicolumn{2}{c}{AP$^\text{mask}$} \\
initialization & data & ViT-B & ViT-L & ViT-B & ViT-L  \\
\shline
supervised  & \scriptsize IN1k w/ labels  & 47.9 & 49.3 & 42.9 & 43.9 \\
\hline
random      & \scriptsize \emph{none}     & 48.9 & 50.7 & 43.6 & 44.9 \\
MoCo v3     & \scriptsize IN1k            & 47.9 & 49.3 & 42.7 & 44.0 \\
BEiT        & \scriptsize IN1k$+$\dalle    & 49.8 & \textbf{53.3} & 44.4 & 47.1 \\
MAE         & \scriptsize IN1k            & \textbf{50.3} & \textbf{53.3} & \textbf{44.9} & \textbf{47.2} \\
\end{tabular}
\vspace{-1em}
\caption{\textbf{COCO object detection and instance segmentation} using our ViT-based Mask R-CNN baseline. Results are reported on COCO 2017 \texttt{val} using the best schedule length (see Figure~\ref{fig:finetuning_epochs}). Random initialization does not use any pre-training data, supervised initialization uses IN1k \emph{with} labels, and all other initializations use IN1k \emph{without} labels. Additionally, BEiT uses a dVAE trained on the proprietary \dalle dataset of \app 250M images~\cite{Ramesh2021}.
\label{tab:main}
}
\vspace{-5mm}
\end{table}
      
\paragraph{Results.} In Table~\ref{tab:main}, we compare COCO fine-tuning results using the pre-trained initializations and random initialization described in~\S\ref{sec:init_methods}. We show results after maximizing \boxAP over the considered training lengths: 25, 50, or 100 epochs for pre-trained initializations, and 100, 200, or 400 epochs for random initialization. (We discuss convergence below.) Next, we make several observations.

\textbf{(1)} Our updated Mask R-CNN trains smoothly with \mbox{ViT-B} and \mbox{ViT-L} backbones regardless of the initialization method. It does not exhibit instabilities nor does it require stabilizing techniques like gradient clipping.

\textbf{(2)} Training from scratch yields up to 1.4 \emph{higher} \boxAP than fine-tuning from supervised IN1k pre-training (50.7 \vs 49.3). While the higher AP may sound surprising, the same trend is observed in~\cite{ghiasi2021simple}. Supervised pre-training is \emph{not} always a stronger baseline than random initialization.

\textbf{(3)} The contrastive learning-based MoCo v3 underperforms random initialization's AP and has similar results compared to supervised initialization.

\textbf{(4)} For ViT-B, BEiT and MAE outperform both random initialization by up to 1.4 \boxAP (50.3 \vs 48.9) and supervised initialization by up to 2.4 \boxAP (50.3 \vs 47.9).

\textbf{(5)}. For ViT-L, the \boxAP gap \emph{increases}, with BEiT and MAE \emph{substantially} outperforming both random initialization by up to 2.6 \boxAP (53.3 \vs 50.7) and supervised initialization by up to 4.0 \boxAP (53.3 \vs 49.3).
  
\begin{figure}[t]\centering
\includegraphics[width=.95\linewidth,trim={0 0 0 0},clip]{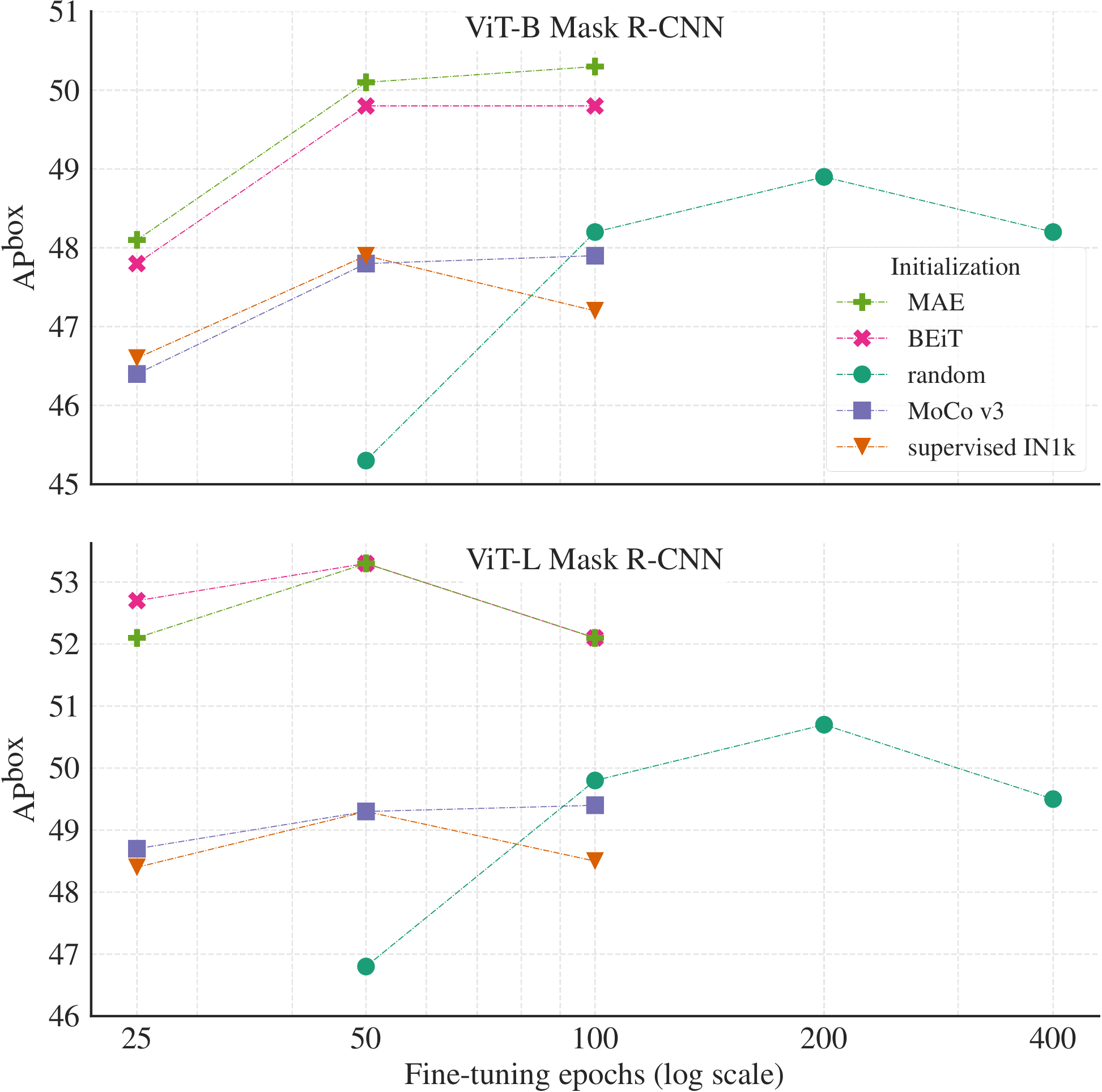}
\vspace{-.5em}
\caption{\textbf{Impact of fine-tuning epochs.} Convergence plots for fine-tuning from 25 and 400 epochs on COCO. All pre-trained initializations converge much faster (\app $4\times$) compared to random initialization, though they achieve varied peak \boxAP. The performance gap between the masking-based methods (MAE and BEiT) and all others is visually evident. When increasing model scale from ViT-B (top) to ViT-L (bottom), this gap also increases, suggesting that these methods may have superior scaling properties.
\label{fig:finetuning_epochs}
}
\end{figure}

\paragraph{Convergence.} In Figure~\ref{fig:finetuning_epochs} we show how pre-training impacts fine-tuning convergence. Given the tuned hyperparameters for each initialization method, we train models for $2\times$ and $4\times$ longer (and also $0.5\times$ for random initialization). Generally, we find that all pre-trained initializations significantly accelerate convergence compared to random initialization, as observed in~\cite{He2019}. Most methods show signs of overfitting when the training schedule is made sufficiently long, typically by 100 epochs for pre-trained initializations and 400 epochs for random initialization. Based on this data, pre-training tends to accelerate training on COCO by roughly $4\times$ compared to random initialization.

We also note two caveats about these results: (i) The drop path rate should ideally be tuned for each training duration as we have observed that the optimal \drp value may need to increase when models are trained for longer. (However, performing an exhaustive \drp sweep for all initializations, model sizes, and training durations is likely computationally impractical.) (ii) Moreover, it may be possible to achieve better results in all cases by training for longer under a more complex training formula that employs heavier regularization and stronger data augmentation.

\paragraph{Discussion.} The COCO dataset is a challenging setting for transfer learning. Due to the large training set (\app 118k images with \app 0.9M annotated objects), it is possible to achieve strong results when training from random initialization. We find that existing methods, like supervised IN1k or unsupervised MoCo v3 pre-training, actually \emph{underperform} the AP of the random initialization baseline (though they yield faster convergence). Prior works reporting unsupervised transfer learning improvements on COCO (\eg, \cite{He2020}) tend to show modest gains over supervised pre-training (\eg, \app 1 \boxAP) and do not include a strong random initialization baseline as we do here (because strong training formulae based on large-scale jitter had not yet been developed). Moreover, they use weaker models and report results that are overall much lower (\eg, \app 40 \boxAP) making it unclear how well the findings translate to state-of-the-art practices.

We find that MAE and BEiT provide the first convincing results of substantial COCO AP improvements due to pre-training. Moreover, these masking-based methods show the potential to improve detection transfer learning as model size increases. We do not observe this important scaling trend with either supervised IN1k pre-training or unsupervised contrastive learning, as represented by MoCo v3.

\subsection{Ablations and Analysis}
\label{sec:ablations}

We ablate several factors involved in the system comparison, analyze model complexity, and report tuned hyperparameter values. For these experiments, we use MAE and 50 epoch fine-tuning by default.

\paragraph{Single-scale \vs Multi-scale.} In Table~\ref{tab:fpn} we compare our default FPN-based multi-scale detector to a single-scale variant. The single-scale variant simply applies RPN and RoIAlign~\cite{He2017} to the final $1/16$th resolution feature map generated by the ViT backbone. The RoI heads and all other choices are the same between the systems (in particular, note that both use the same hybrid windowed/global attention). We observe that the multi-scale FPN design increases \boxAP by \app 1.3-1.7 (\eg, 50.1 \vs 48.4), while increasing training and inference time by \app 5 and \app 10\% relative, respectively. Multi-scale memory overhead is $<$1\%.

\begin{table}
\tablestyle{6pt}{1.1}
\begin{tabular}{@{}lcc@{}}
& \multicolumn{2}{c}{AP$^\text{box}$} \\
FPN & ViT-B & ViT-L \\
\shline
\baseline{yes} & \textbf{50.1} & \textbf{53.3} \\
no  & 48.4 & 52.0 \\
\end{tabular}
\vspace{-1em}
\caption{\textbf{Single-scale \vs multi-scale (FPN) ablation.} FPN yields consistent improvements. Our default setting is marked in gray.
\label{tab:fpn}
}
\end{table}

\paragraph{Memory and Time Reduction.} In Table~\ref{tab:win_attn} we compare several strategies for reducing memory and time complexity when using a standard ViT backbone in Mask R-CNN. Using a combination of $14\times 14$ non-overlapping windowed self-attention together with four global attention blocks achieves a good balance between memory, training and inference time, and AP metrics. This finding motivates us to use this setting as our default. Somewhat surprisingly using only windowed attention is \emph{not} catastrophic even though the backbone processes all windows entirely independently (\boxAP decreases from 53.3 to 50.7). This is likely due to cross-window computation introduced by convolutions and RoIAlign in the rest of the Mask R-CNN model.

\begin{table}
\tablestyle{6pt}{1.1}
\begin{tabular}{@{}b{2.6cm}b{0.6cm}cb{0.6cm}b{0.6cm}b{0.6cm}@{}}
self-attention & act checkpt & AP$^\text{box}$ & train mem & train time & test time \\
\shline
(1) windowed                      & no  & 50.7 & 16GB & \textbf{0.67s} & \textbf{0.34s} \\
(2) \baseline{windowed, 4 global} & \baseline{no}  & \textbf{53.3} & 27GB & 0.93s & 0.40s \\
(3) global                        & yes & 53.1 & \textbf{14GB} & 2.26s & 0.65s \\
(4) global                        & no  & - & OOM & - & - \\
\end{tabular}
\vspace{-1em}
\caption{\textbf{Memory and time reduction strategies.} We compare methods for reducing memory and time when using ViT-L in Mask R-CNN. The strategies include: (1) replace \emph{all} global self-attention with $14\times 14$ non-overlapping windowed self-attention, (2) a hybrid that uses both windowed and global self-attention, or (3) all global attention with activation checkpointing. Without any of these strategies (row 4) an out-of-memory (OOM) error prevents training. We report \boxAP, peak GPU training memory, average per-iteration training time, and average per-image inference time using NVIDIA V100-32GB GPUs. The per-GPU batch size is 1. Our defaults (row 2) achieves a good balance between memory, time, and \boxAP metrics. In fact, our hybrid approach achieves comparable \boxAP to full global attention, while being much faster.
\label{tab:win_attn}
}
\end{table}

\paragraph{Positional Information.} In the default BEiT code, the ViT is modified to use relative position biases~\cite{raffel2019exploring} in each transformer block instead of adding absolute position embeddings to the patch embeddings. This choice is an orthogonal enhancement that is not used by the other pre-training methods (though it could be). In an attempt to make the comparison more equal, we include these biases (and absolute position embeddings) in all fine-tuning models by default, as discussed in \S\ref{sec:nuisance}.

In Table~\ref{tab:rel_pos} we study the effect of relative position biases on fine-tuning performance. A detailed analysis is given in the caption. In summary, we observe that including relative position biases during fine-tuning may slightly improve \boxAP by \app 0.2--0.3 points (\eg, 53.0 to 53.3) for a model that was pre-trained with only absolute position embeddings. We also observe that pre-training relative position biases, as done by BEiT, may also have a slight positive effect of \app 0.1--0.3 points. Our practice of including both positional information types during fine-tuning appears to provide a reasonably fair comparison. We also note that using relative position biases introduces non-trivial overhead---it increases training and inference time by roughly 25\% and 15\% relative, respectively, increases memory by \app 15\% (even with shared biases), and perhaps should be avoided.

\begin{table}[t!]
\tablestyle{6pt}{1.1}
\begin{tabular}{@{}lcccccc@{}}
                & \multicolumn{2}{c}{\scriptsize pre-train (pt)} & \multicolumn{2}{c}{\scriptsize fine-tuning} & \multicolumn{2}{c}{\boxAP} \\
initialization & abs & rel                          & abs  & rel                          & ViT-B & ViT-L \\
\shline
(1) BEiT       & \baseline{no} & \baseline{yes}     & \baseline{rand} & \baseline{pt}     & 49.8  & \textbf{53.3} \\
(2) BEiT       & no  & yes                          & rand & zero                         & 49.5  & 53.2 \\
\hline
(3) BEiT$^\dagger$       & yes & no                           & pt   & zero                         & -     & 53.1 \\
\hline
(4) MAE        & \baseline{yes} & \baseline{no}     & \baseline{pt} & \baseline{zero}     & \textbf{50.1}  & \textbf{53.3} \\
(5) MAE        & yes & no                           & pt   & no                           & 49.9  & 53.0 \\
\end{tabular}
\vspace{-1em}
\caption{\textbf{Positional information ablation.} In the BEiT code, the ViT is modified to use relative position biases (\emph{rel}) instead of absolute position embeddings (\emph{abs}). We study how these components impact results based on their use in pre-training (\emph{pt}) and under various treatments in fine-tuning: (i) \emph{pt}: initialized with pre-trained values; (ii) \emph{rand}: random initialization; (iii) \emph{zero}: initialized at zero; and (iv) \emph{no}: this positional information is not used in the fine-tuned model. For BEiT$^\dagger$ (row 3), we pre-train an additional model (ViT-L only) that, like MAE, uses absolute position embeddings instead of relative position biases. Our default settings are marked in gray. Comparing (1) and (2), we observe that pre-trained relative position bias initialization provides a slight benefit over zero initialization. Comparing (1,2) to (3), we see that BEiT pre-trained with absolute position embeddings performs similarly (perhaps slightly worse) to pre-training with relative position biases. Comparing (4) and (5), we see that including relative position biases in addition to absolute position embeddings provides a small improvement.
\label{tab:rel_pos}
}
\end{table}
 
\begin{figure}[t]\centering
\includegraphics[width=.95\linewidth,trim={0 0 0 0},clip]{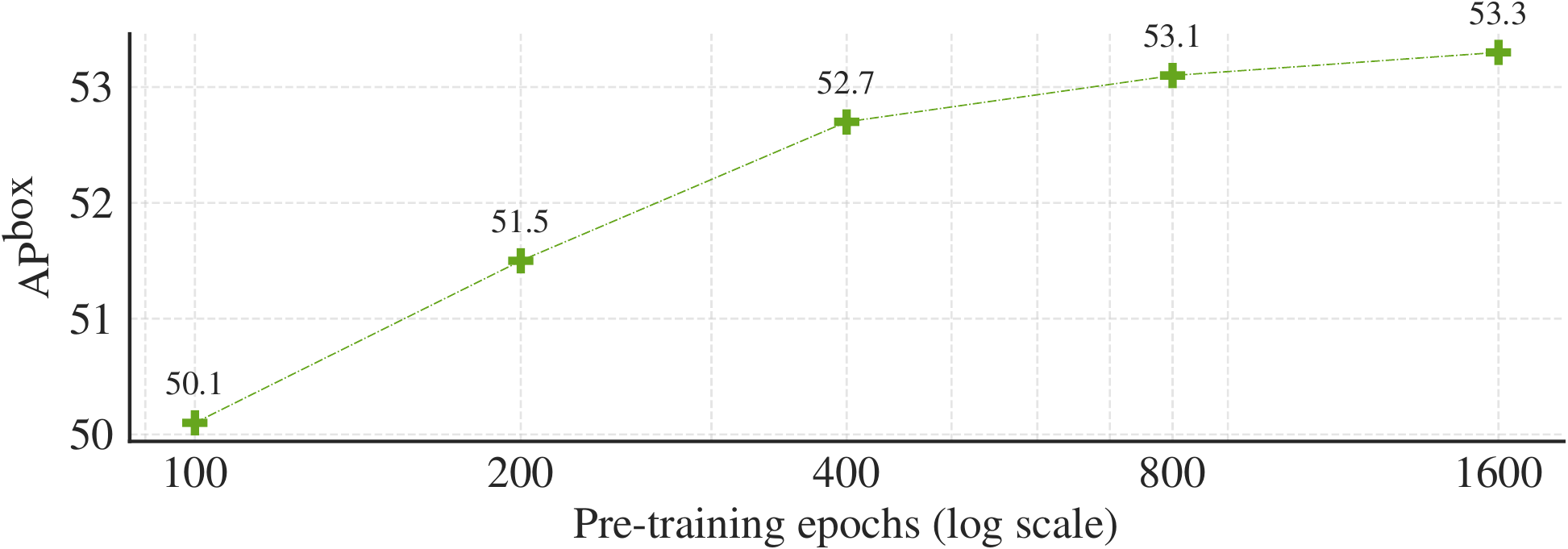}
\vspace{-.5em}
\caption{\textbf{Impact of pre-training epochs.} Increasing MAE pre-training from 100 to 800 epochs confers large transfer learning gains. The improvements start to plateau after 800 epochs.
\label{fig:pre_training_epochs}
}
\end{figure}

\paragraph{Pre-training Epochs.} In Figure~\ref{fig:pre_training_epochs} we study the impact of MAE pre-training epochs on COCO \boxAP by sweeping pre-training epochs from 100 to 1600 (the default). The results show that pre-training duration has a significant impact on transfer learning performance with large increases in \boxAP continuing from 100 to 800 epochs. There is still a small improvement from 800 to 1600 epochs ($+$0.2 from 53.1 to 53.3), though the gradient has largely flattened.

\begin{figure}[t]\centering
\includegraphics[width=.95\linewidth,trim={0 0 0 0},clip]{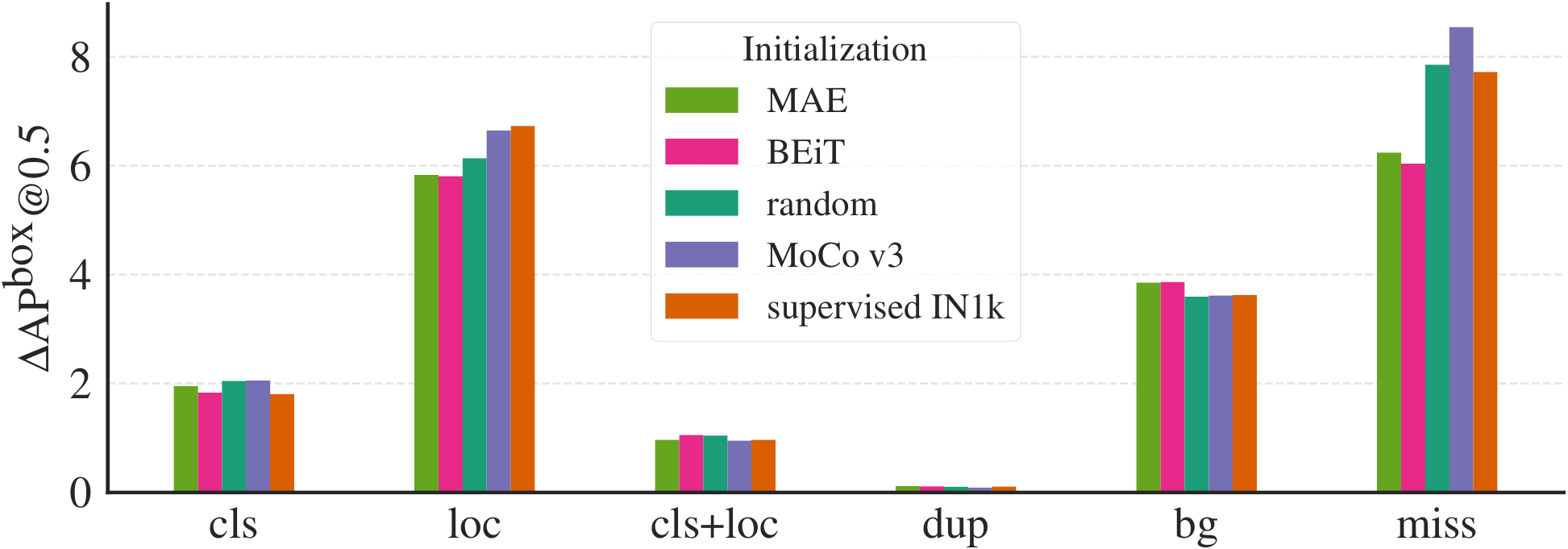}
\vspace{-.5em}
\caption{\textbf{TIDE analysis.} We plot the $\Delta$\boxAP metric at an intersection-over-union (IoU) threshold of 0.5 as defined in~\cite{bolya2020tide}. Each bar shows how much AP can be added to the detector if an oracle fixes a certain error type. The error types are: \emph{cls}: localized correctly (IoU $\ge$0.5), but classified incorrectly; \emph{loc}: classified correctly, but localized incorrectly (IoU in $[$0.1, 0.5$)$); \emph{cls+loc}: classified incorrectly and localized incorrectly; \emph{dup}: detection would be correct if not for a higher scoring correct detection; \emph{bg}: detection is in the background (IoU $<$0.1); \emph{miss}: all undetected ground-truth objects not covered by other error types. (See~\cite{bolya2020tide} for more details and discussion.) We observe that the masking-based initializations (MAE and BEiT) make fewer localization errors than MoCo v3 and supervised initialization (random initialization is somewhere in-between) and, even more so, have fewer missed detections. The other error types are more similar across initializations.
\label{fig:tide}
}
\vspace{-1mm}
\end{figure}

\paragraph{TIDE Error Type Analysis.} In Figure~\ref{fig:tide} we show the error type analysis generated by the TIDE toolbox~\cite{bolya2020tide}. A detailed description and analysis is given in the caption. The analysis reveals more granular information about \emph{where} MAE and BEiT improve overall AP relative to the other initializations. In summary, we observe that all initializations lead to roughly the same classification performance for correctly localized objects, however the MAE and BEiT initializations improve localization compared to the other initializations. We observe an even stronger effect when looking at missed detections: the masking-based initializations yield notably higher recall than the other initializations and thus leave fewer undetected objects. This higher recall creates a small increase in background errors, thus leading to better overall AP.

\begin{table}
\tablestyle{6pt}{1.1}
\begin{tabular}{@{}lcccc@{}}
backbone & params (M) & acts (M) & flops (G) & fps \\
\shline
ResNet-101  & \phantom{0}65 & \phantom{0}426 {\scriptsize $\pm$ 43} & \phantom{0}422 {\scriptsize $\pm$ 35} & 13.7 \\
\hline
ViT-B       & 116 & 1532 {\scriptsize $\pm$ 11} & \phantom{0}853 {\scriptsize $\pm$ 13} & \phantom{0}5.1 \\
ViT-L       & 339 & 2727 {\scriptsize $\pm$ 10} & 1907 {\scriptsize $\pm$ 12} & \phantom{0}2.5 \\
\end{tabular}
\vspace{-1em}
\caption{\textbf{Model complexity for inference} with the specific Mask R-CNN configuration used in this report. For ViT, the image resolution is $1024\times 1024$ (padded as necessary). The flop and activation counts are measured at runtime and vary based on the number of detected objects. We report the mean $\pm$ one standard deviation from 100 validation images. Results change very slightly when using different initializations. For reference, we report results using the ResNet-101 backbone, which can (and does) use non-square inputs at inference time (longest side is $1024$); otherwise inference settings are the same. The ResNet-101 based Mask R-CNN achieves 48.9 \boxAP when trained from scratch for 400 epochs. We also report wall-clock speed in frames-per-second (fps) on an NVIDIA V100-32GB GPU.
\label{tab:complexity}
}
\end{table}

\paragraph{Model Complexity.} Table~\ref{tab:complexity} compares various complexity and wall-clock time measures of our specific Mask R-CNN configuration. We also report these measures using a ResNet-101 backbone instead of ViT. When trained from scratch, both ResNet-101 and ViT-B backbones achieve 48.9 \boxAP. At inference time, the ResNet-101 backbone is much faster; however, during training ViT-B reaches peak performance at 200 epochs compared to 400 for ResNet-101. ResNet-101 is not yet able to benefit from BEiT or MAE pre-training and therefore lags behind ViT-B in \boxAP (\app 1 point) when those methods are used for initialization.

\paragraph{Hyperparameter Tuning.} All pre-trained initializations preferred \wtd $=0.1$ for fine-tuning. Random initialization benefitted from stronger regularization and selected a higher setting of $0.2$. Most methods selected \lr $= \expnum{8.0}{-5}$, except for random initialization and MoCo v3 initialization, which both preferred a higher setting of $\expnum{1.6}{-4}$. As described previously, the drop path rate could not be reliably tuned using shorter schedules. As a result, we tuned \drp with 50 epoch training for pre-trained initializations and 100 epoch training for random initialization. Based on this tuning, all initializations selected \drp $=0.1$ when using ViT-B and $0.3$ when using ViT-L.


\section{Conclusion}
\label{sec:conclusion}
We have presented techniques that enable the practical use of standard ViT models as the backbone in Mask \mbox{R-CNN}. These methods yield acceptable training memory and time, while also achieving strong results on COCO without involving too many complex extensions. Using these techniques, we find effective training formulae that enable us to benchmark five different ViT initialization methods. We show that random initialization takes \app $4\times$ longer than any of the pre-trained initializations, but achieves a meaningfully higher AP than ImageNet-1k supervised pre-training. We find that MoCo v3, a representative of contrastive unsupervised learning, performs nearly the same as supervised pre-training (and thus worse than random initialization). Importantly, we witness an exciting new result: masking-based methods (BEiT and MAE) show considerable gains over both supervised and random initialization and these gains increase as model size increases. This scaling behavior is not observed with either supervised or MoCo v3-based initialization.

{\small
\bibliographystyle{ieee_fullname}
\bibliography{vit_det}

\begin{thebibliography}{10}\itemsep=-1pt

\bibitem{Bao2021}
Hangbo Bao, Li Dong, and Furu Wei.
\newblock {BEiT}: Bert pre-training of image transformers.
\newblock {\em arXiv:2106.08254}, 2021.

\bibitem{beal2020toward}
Josh Beal, Eric Kim, Eric Tzeng, Dong~Huk Park, Andrew Zhai, and Dmitry
  Kislyuk.
\newblock Toward transformer-based object detection.
\newblock {\em arXiv preprint arXiv:2012.09958}, 2020.

\bibitem{bolya2020tide}
Daniel Bolya, Sean Foley, James Hays, and Judy Hoffman.
\newblock {TIDE}: A general toolbox for identifying object detection errors.
\newblock In {\em ECCV}, 2020.

\bibitem{Cai2018}
Zhaowei Cai and Nuno Vasconcelos.
\newblock {Cascade R-CNN}: Delving into high quality object detection.
\newblock In {\em CVPR}, 2018.

\bibitem{Caron2021}
Mathilde Caron, Hugo Touvron, Ishan Misra, Herv{\'e} J{\'e}gou, Julien Mairal,
  Piotr Bojanowski, and Armand Joulin.
\newblock Emerging properties in self-supervised vision transformers.
\newblock In {\em ICCV}, 2021.

\bibitem{chen2019hybrid}
Kai Chen, Jiangmiao Pang, Jiaqi Wang, Yu Xiong, Xiaoxiao Li, Shuyang Sun,
  Wansen Feng, Ziwei Liu, Jianping Shi, Wanli Ouyang, et~al.
\newblock Hybrid task cascade for instance segmentation.
\newblock In {\em CVPR}, 2019.

\bibitem{Chen2021a}
Xinlei Chen, Saining Xie, and Kaiming He.
\newblock An empirical study of training self-supervised {Vision Transformers}.
\newblock In {\em ICCV}, 2021.

\bibitem{Deng2009}
Jia Deng, Wei Dong, Richard Socher, Li-Jia Li, Kai Li, and Li Fei-Fei.
\newblock {ImageNet: A large-scale hierarchical image database}.
\newblock In {\em CVPR}, 2009.

\bibitem{Doersch2015}
Carl Doersch, Abhinav Gupta, and Alexei~A Efros.
\newblock Unsupervised visual representation learning by context prediction.
\newblock In {\em ICCV}, 2015.

\bibitem{Dosovitskiy2021}
Alexey Dosovitskiy, Lucas Beyer, Alexander Kolesnikov, Dirk Weissenborn,
  Xiaohua Zhai, Thomas Unterthiner, Mostafa Dehghani, Matthias Minderer, Georg
  Heigold, Sylvain Gelly, Jakob Uszkoreit, and Neil Houlsby.
\newblock An image is worth 16x16 words: Transformers for image recognition at
  scale.
\newblock In {\em ICLR}, 2021.

\bibitem{el2021xcit}
Alaaeldin El-Nouby, Hugo Touvron, Mathilde Caron, Piotr Bojanowski, Matthijs
  Douze, Armand Joulin, Ivan Laptev, Natalia Neverova, Gabriel Synnaeve, Jakob
  Verbeek, et~al.
\newblock {XCiT}: Cross-covariance image transformers.
\newblock {\em arXiv preprint arXiv:2106.09681}, 2021.

\bibitem{fan2021multiscale}
Haoqi Fan, Bo Xiong, Karttikeya Mangalam, Yanghao Li, Zhicheng Yan, Jitendra
  Malik, and Christoph Feichtenhofer.
\newblock Multiscale vision transformers.
\newblock {\em arXiv preprint arXiv:2104.11227}, 2021.

\bibitem{ghiasi2021simple}
Golnaz Ghiasi, Yin Cui, Aravind Srinivas, Rui Qian, Tsung-Yi Lin, Ekin~D Cubuk,
  Quoc~V Le, and Barret Zoph.
\newblock Simple copy-paste is a strong data augmentation method for instance
  segmentation.
\newblock In {\em CVPR}, 2021.

\bibitem{Girshick2014}
Ross Girshick, Jeff Donahue, Trevor Darrell, and Jitendra Malik.
\newblock Rich feature hierarchies for accurate object detection and semantic
  segmentation.
\newblock In {\em CVPR}, 2014.

\bibitem{Goyal2017}
Priya Goyal, Piotr Doll{\'a}r, Ross Girshick, Pieter Noordhuis, Lukasz
  Wesolowski, Aapo Kyrola, Andrew Tulloch, Yangqing Jia, and Kaiming He.
\newblock Accurate, large minibatch {SGD}: Training {ImageNet} in 1 hour.
\newblock {\em arXiv:1706.02677}, 2017.

\bibitem{he2021mae}
Kaiming He, Xinlei Chen, Saining Xie, Yanghao Li, Piotr Doll{\'a}r, and Ross
  Girshick.
\newblock Masked autoencoders are scalable vision learners.
\newblock {\em arXiv preprint arXiv:2111.06377}, 2021.

\bibitem{He2020}
Kaiming He, Haoqi Fan, Yuxin Wu, Saining Xie, and Ross Girshick.
\newblock Momentum contrast for unsupervised visual representation learning.
\newblock In {\em CVPR}, 2020.

\bibitem{He2019}
Kaiming He, Ross Girshick, and Piotr Doll{\'a}r.
\newblock Rethinking {ImageNet} pre-training.
\newblock In {\em ICCV}, 2019.

\bibitem{He2017}
Kaiming He, Georgia Gkioxari, Piotr Doll{\'a}r, and Ross Girshick.
\newblock {Mask R-CNN}.
\newblock In {\em ICCV}, 2017.

\bibitem{He2016}
Kaiming He, Xiangyu Zhang, Shaoqing Ren, and Jian Sun.
\newblock Deep residual learning for image recognition.
\newblock In {\em CVPR}, 2016.

\bibitem{Hendrycks2016}
Dan Hendrycks and Kevin Gimpel.
\newblock Gaussian error linear units (gelus).
\newblock {\em arXiv:1606.08415}, 2016.

\bibitem{Huang2016}
Gao Huang, Yu Sun, Zhuang Liu, Daniel Sedra, and Kilian~Q Weinberger.
\newblock Deep networks with stochastic depth.
\newblock In {\em ECCV}, 2016.

\bibitem{Ioffe2015}
Sergey Ioffe and Christian Szegedy.
\newblock Batch normalization: Accelerating deep network training by reducing
  internal covariate shift.
\newblock In {\em ICML}, 2015.

\bibitem{Larsson2016}
Gustav Larsson, Michael Maire, and Gregory Shakhnarovich.
\newblock Fractalnet: Ultra-deep neural networks without residuals.
\newblock {\em ICLR}, 2016.

\bibitem{LeCun1989}
Yann LeCun, Bernhard Boser, John~S Denker, Donnie Henderson, Richard~E Howard,
  Wayne Hubbard, and Lawrence~D Jackel.
\newblock Backpropagation applied to handwritten zip code recognition.
\newblock {\em Neural computation}, 1989.

\bibitem{mvitv2}
Yanghao Li, Chao-Yuan Wu, Haoqi Fan, Karttikeya Mangalam, Bo Xiong, Jitendra
  Malik, and Christoph Feichtenhofer.
\newblock Improved multiscale vision transformers for classification and
  detection.
\newblock {\em In preparation}, 2021.

\bibitem{Lin2017}
Tsung-Yi Lin, Piotr Doll{\'a}r, Ross Girshick, Kaiming He, Bharath Hariharan,
  and Serge Belongie.
\newblock Feature pyramid networks for object detection.
\newblock In {\em CVPR}, 2017.

\bibitem{Lin2014}
Tsung-Yi Lin, Michael Maire, Serge Belongie, James Hays, Pietro Perona, Deva
  Ramanan, Piotr Doll{\'a}r, and C~Lawrence Zitnick.
\newblock {Microsoft COCO: Common objects in context}.
\newblock In {\em ECCV}. 2014.

\bibitem{liu2021swin}
Ze Liu, Yutong Lin, Yue Cao, Han Hu, Yixuan Wei, Zheng Zhang, Stephen Lin, and
  Baining Guo.
\newblock Swin transformer: Hierarchical vision transformer using shifted
  windows.
\newblock {\em arXiv preprint arXiv:2103.14030}, 2021.

\bibitem{Loshchilov2019}
Ilya Loshchilov and Frank Hutter.
\newblock Decoupled weight decay regularization.
\newblock In {\em ICLR}, 2019.

\bibitem{raffel2019exploring}
Colin Raffel, Noam Shazeer, Adam Roberts, Katherine Lee, Sharan Narang, Michael
  Matena, Yanqi Zhou, Wei Li, and Peter~J Liu.
\newblock Exploring the limits of transfer learning with a unified text-to-text
  transformer.
\newblock {\em arXiv preprint arXiv:1910.10683}, 2019.

\bibitem{Ramesh2021}
Aditya Ramesh, Mikhail Pavlov, Gabriel Goh, Scott Gray, Chelsea Voss, Alec
  Radford, Mark Chen, and Ilya Sutskever.
\newblock Zero-shot text-to-image generation.
\newblock {\em arXiv:2102.12092}, 2021.

\bibitem{Ren2015}
Shaoqing Ren, Kaiming He, Ross Girshick, and Jian Sun.
\newblock {Faster R-CNN}: Towards real-time object detection with region
  proposal networks.
\newblock In {\em NeurIPS}, 2015.

\bibitem{Ren2017}
Shaoqing Ren, Kaiming He, Ross Girshick, and Jian Sun.
\newblock {Faster R-CNN}: Towards real-time object detection with region
  proposal networks.
\newblock {\em TPAMI}, 2017.

\bibitem{Sermanet2013}
Pierre Sermanet, Koray Kavukcuoglu, Sandhya Chintala, and Yann LeCun.
\newblock Pedestrian detection with unsupervised multi-stage feature learning.
\newblock In {\em CVPR}, 2013.

\bibitem{Touvron2020}
Hugo Touvron, Matthieu Cord, Matthijs Douze, Francisco Massa, Alexandre
  Sablayrolles, and Herv{\'e} J{\'e}gou.
\newblock Training data-efficient image transformers \& distillation through
  attention.
\newblock {\em arXiv:2012.12877}, 2020.

\bibitem{touvron2021going}
Hugo Touvron, Matthieu Cord, Alexandre Sablayrolles, Gabriel Synnaeve, and
  Herv{\'e} J{\'e}gou.
\newblock Going deeper with image transformers.
\newblock {\em arXiv preprint arXiv:2103.17239}, 2021.

\bibitem{Vaswani2017}
Ashish Vaswani, Noam Shazeer, Niki Parmar, Jakob Uszkoreit, Llion Jones,
  Aidan~N Gomez, Lukasz Kaiser, and Illia Polosukhin.
\newblock Attention is all you need.
\newblock In {\em NeurIPS}, 2017.

\bibitem{Wu2018}
Yuxin Wu and Kaiming He.
\newblock Group normalization.
\newblock In {\em ECCV}, 2018.

\bibitem{Wu2019}
Yuxin Wu, Alexander Kirillov, Francisco Massa, Wan-Yen Lo, and Ross Girshick.
\newblock Detectron2.
\newblock \url{https://github.com/facebookresearch/detectron2}, 2019.

\end{thebibliography}
}

\end{document}